# SODA: Site Object Detection dAtaset for Deep Learning in Construction

**Rui Duan[a], Hui Deng[a], Mao Tian[b], Yichuan Deng[a,c*], Jiarui Lin[d]**

[a] School of Civil Engineering and Transportation, South China University of Technology, Guangzhou, China.

[b] Sonny Astani Department of Civil and Environmental Engineering, Viterbi School of Engineering, University of Southern California, Los Angeles, USA.

[c] State Key Laboratory of Subtropical Building Science, Guangzhou 510641, China.

[d] Department of Civil Engineering, Tsinghua University, Beijing 10084, China.

[*] Corresponding author at: School of Civil Engineering and Transportation, South China University of Technology, Guangzhou 510641, China. E-mail address: ctycdeng@scut.edu.cn.

**Abstract:** Computer vision-based deep learning object detection algorithms have been developed sufficiently powerful to support the ability to recognize various objects. Although there are currently general datasets for object detection, there is still a lack of large-scale, open-source dataset for the construction industry, which limits the developments of object detection algorithms as they tend to be data-hungry. Therefore, this paper develops a new large-scale image dataset specifically collected and annotated for the construction site, called Site Object Detection dAtaset (SODA), which contains 15 kinds of object classes categorized by workers, materials, machines, and layout. Firstly, more than 20,000 images were collected from multiple construction sites in different site conditions, weather conditions, and construction phases, which covered different angles and perspectives. After careful screening and processing, 19,846 images including 286,201 objects were then obtained and annotated with labels in accordance with predefined categories. Statistical analysis shows that the developed dataset is advantageous in terms of diversity and volume. Further evaluation with two widely-adopted object detection algorithms based on deep learning (YOLO v3/ YOLO v4) also illustrates the feasibility of the dataset for typical construction scenarios, achieving a maximum mAP of 81.47%. In this manner, this research contributes a large-scale image dataset for the development of deep learning-based object detection methods in the construction industry and sets up a performance benchmark for further evaluation of corresponding algorithms in this area.

## 1. Introduction

The construction industry is still a labor-intensive industry, with most management and interventions of on-site activities relying on manual judgments [1], which makes construction site management difficult and inefficient. Although the emergence of high-resolution monitoring cameras makes remote and dynamic monitoring of the construction site possible, it still requires a lot of manual intervention [2]. The rapid development of computer vision technology makes it possible to automate tasks that cannot be completed by the human vision system, effectively improving safety and production efficiency [3]. The importance of cameras in the field of construction management has become increasingly prominent, and practitioners have begun to embrace changes brought by automated applications powered by computer vision [4]. For example, video surveillance can identify workers' unsafe behaviors and risks of construction [5], where computer vision technology is used to identify workers who do not wear personal protective equipment [6-11]. The use of computer vision technology in construction automation has thus attracted wide attention from academia and industry.

In recent years, deep learning object detection algorithm has developed rapidly, and many object detection algorithms have emerged. The detection speed and accuracy have been greatly improved. Under the appropriate application scenarios, the recognition accuracy can reach 98% or even higher. At the same time, the detection accuracy of the computer vision technology based on deep learning has great advantages over the traditional image processing and recognition methods [12], and it is far superior to the traditional image processing methods in terms of detection speed, algorithm robustness and feature extraction without manual design. Therefore, the introduction of the deep learning method into object detection in construction site management will be a new direction [13]. However, deep learning algorithms are data-hungry, which means the application of deep learning object detection on construction site requires a specific image dataset in the construction field. Construction is a highly professional process with unique processes that brings challenges to both the collection and annotation of the images, which is the reason that well-annotated image sets for the construction industry are hardly seen in popular image sets such as the ImageNet.

In order to promote the research of object detection in the construction industry, it is necessary to build a large-scale image dataset containing specific objects from the construction site (i.e., workers, materials, machines, layouts). The existing construction site image dataset is relatively small and has fewer categories, concentrating on people, personal protective equipment (PPE), and some machines. This is because: (1) The image of the construction site is more challenging to obtain than that of ordinary objects. Due to security concerns, the construction site is generally not open to the public. Moreover, the available online resources of construction site images are less common than daily objects and have high repeatability. (2) It is difficult to obtain data from different perspectives of objects on construction sites by using the conventional monocular camera installed on-site, which is easy to cause overfitting of object detection model. (3) The environment of the site is usually disorderly and numerous, and the difficulty and cost of annotation are high. It takes professional knowledge to correctly annotate the objects in the images taken from the construction site.

A general object detection image dataset for construction sites will benefit the construction industry in terms of serving as the basis for generating deep-learning-based object detection models and testing object detection algorithms. Considering the professional knowledge involved in building such a dataset, it remains a task to be resolved by people in the industry. The purpose of this research is thus to construct a general object detection image dataset for the construction site. The research processes include category selection, data acquisition, data cleaning, data annotation, dataset analysis, and experimental analysis and benchmark. A dataset of 19,846 images including 286201 objects with 15 categories of tags was constructed manually in this process. More than 20,000 visual image data of different construction sites were collected using various equipment including monocular camera, UAV, hook visualization equipment. Then 35 students majoring in Civil Engineering were recruited and trained to process and annotate the data, and the authors conducted data statistics on the developed dataset. Finally, the dataset is tested in the mainstream object detection algorithms, with satisfactory verification results obtained, which provides datasets and corresponding algorithm benchmarks for subsequent research. The images and annotations were then released online, allowing unlimited public access. In the future, it will be iteratively upgraded regularly, improving the variety of construction site objects and increasing more image data to promote the development of automatic construction.

The remaining of article is structured as follows: the second section of this paper introduces the work of object detection and related image datasets. The third section introduces the process of data acquisition and annotation. In the fourth section, we make the statistics of the dataset. The fifth section introduces the experimental results of two mainstream one-stage object detection algorithms on our dataset, which provides the benchmark for subsequent research. The contributions of this study are as follows: (1)19846 construction site images containing 15 classes of objects and VOC format dataset are built and publicly provided for the construction industry. SODA not only provides Image data with considerable quantity and quality but also includes four categories: workers, materials, machines, and layout for the first time. (2) It provides a markup of production logic and shares relevant experience in building such data sets, according to which we can increase the number of categories and images in the future. (3) It provides a benchmark for object detection using object detection algorithms in the construction industry.

## 2. Related Work

### *2.1 Computer vision and deep learning in construction*

The use of computer vision technology for construction has aroused strong interest in academia and industry. At present, there have been many applications of computer vision in the construction industry, such as safety monitoring, progress productivity analysis, and personnel management. Koch et al. [14] reviewed the application of computer vision technology in defect detection and condition assessment of concrete and asphalt civil infrastructure. Xiang et al. [15] proposed an intelligent monitoring method based on deep learning for locating and identifying intrusion engineering vehicles, which can prevent the external damage of buildings. In order to prevent construction workers from falling from high, Fang et al. [16] developed an automatic method based on two convolutional neural networks (CNN) models to determine whether workers wear safety belts at high altitudes. Yang et al. [17] used tower crane cameras to record video data and used MASK R-CNN to identify pixel coordinates of workers and dangerous areas, and then applied it to the management of field staff and hazard sources. Chen et al. [18] used construction site surveillance videos to detect, track and identify excavator activities in the framework using three different convolutional neural networks. According to the results of convolutional neural network recognition, the excavator activity time, working cycle, and productivity are analyzed. Deng et al. [19] proposed a method combining computer vision with building information modeling (BIM) to realize automatic progress monitoring of tile installation. This method can automatically and accurately measure the construction progress information at the construction site. Fang et al. [20] realized the real-time positioning of construction-related entities by using deep learning visual algorithms combined with semantic and prior knowledge. Zhang et al. [21] proposed an automatic identification method based on the combination of deep learning object detection and ontology. This method can effectively identify the risks in the construction process and prevent the occurrence of construction accidents. Luo et al. [22] use computer vision and deep learning technology to automatically estimate the posture of different construction equipment in the video taken at the construction site. Pan et al. [23] proposed a novel framework named Video2entities, which combines visual data and common knowledge graph as a priori information and uses zero-shot learning (ZSL) to realize the detection of unknown targets, and effectively improve the self-learning ability of the object detection algorithm. Computer vision is changing the construction process, so this study collected some reviews [24-26] of computer vision in the construction industry for the reference of interested readers. They also noticed that the lack of sufficient scale of image database and data privacy-related issues would hinder the continuous development of computer vision in the construction field. They agreed that defect inspection, safety monitoring, and performance analysis of computer vision in the construction field remain a rich exploration direction.

### *2.2 Deep Learning Object Detection Datasets*

According to the domain coverage, the existing computer vision dataset can be divided into two categories, general image dataset, and domain image dataset. General datasets can be used by the public, including natural categories in daily life. Compared with the general dataset, the domain dataset contains the categories of specific fields.

General datasets are mainly for natural categories, such as people, animals, vehicles, etc. The Mnist dataset [27] contains 70,000 handwritten digital images of 28 * 28 size, which is an entry-level dataset for deep learning. The PASCAL VOC dataset [28], which is widely used in testing deep learning algorithms contains 11,000 images of 20 classes, with a size of about 2GB. Microsoft COCO [29] is a dataset with 160,000 images developed by Microsoft with 91 categories. It has a text description of the category, location, and image, and is about 40GB in size. ImageNet [30] was created by Professor Li Feifei’s team. It has more than 14 million images, covering more than 20,000 categories and about 1TB in size. Most of the research work on image classification, location, and detection benefits from this dataset. The capacity and types of general datasets are also constantly increasing and improving, giving birth to a large number of excellent computer vision models (especially those based on deep learning), and promoting the rapid development of computer vision.

The general image dataset introduced above often lacks most categories in the field of construction. In recent years, some image datasets have also appeared in the field of building construction. Tajudeen et al. [31] collected thousands of images of construction machines and created an image dataset covering five kinds of construction equipment (excavators, loaders, bulldozers, rollers, and backhoe diggers). Kolar et al. [32] focused on construction site guardrail detection, adding background images to the three-dimensional model of the guardrail, creating an enhanced dataset containing 6,000 images, and performing image detection with VGG16 and CNN. Li et al. [6] established and released a dataset containing 3261 images of safety helmet, and used the SSD-MobileNet algorithm based on a convolutional neural network to detect the unsafe operation of wearing helmets on construction sites. An et al. [33] collected more than 40,000 images from 174 construction sites to annotate 13 types of moving objects and created the ' Moving Objects in Construction Site ' (MOCS) dataset. They used pixel segmentation to annotate objects precisely and tested them on 15 different deep neural networks. Wang et al. [7] constructed a dedicated Personal Protective Equipment (PPE) image dataset Color Helmet and Vest (CHV) composed of 1,330 images. They used eight YOLO algorithm detectors to test and compare performance on CHV datasets. Xiao et al. [34] developed the Alberta Construction Image Dataset (ACID), an image dataset specially used to identify construction machinery, and manually collected and marked 10,000 images of 10 kinds of construction machines. Four object detection algorithms, YOLO v3, Inception SSD, R-FCN-ResNet101, and Faster-RCNN-ResNet101, were used to train the dataset, and better mAP indicators and average detection speed were obtained.

As mentioned above, visual data in the field of construction is also growing. But at present, there are few datasets in the field of construction, which are often concentrated in construction workers, PPE, and some machines. However, we also have the necessity of object detection for the materials and construction site layout of the construction site. Identifying the location of construction site materials can not only improve labor performance but also avoid risks of construction. For example, monitoring the collision problem of the tower crane during hoisting. Identifying the location of the object in the construction site can not only judge the safety of the construction scene through a series of future works but also judge whether the construction process is performed in a civilized manner. According to the literature review, there are few studies on the identification and detection of construction materials [35,36] and layout in the construction industry. Therefore, it is necessary to further develop image datasets containing workers, machines, material, and layouts for further research on deep learning in the construction industry.

### *2.3 Deep Learning Object Detection Algorithms*

Object detection is an important direction of deep learning, which focuses on specific object targets and obtains the category information and location information of the object target [35]. At present, the deep learning object detection algorithm is divided into two-stage object detection and one-stage object detection. The two-stage model is also named as region-based method because of its two-stage processing of images. It extracts a series of checkboxes and then uses convolutional neural networks for classification. Two-stage object detection has higher precision but relatively slow speed. Representative algorithms such as R-CNN [37], Fast R-CNN [38], Faster R-CNN [39]; YOLO [40] has created a one-stage object detection algorithm, which directly obtains the prediction results from the image, and is also called the end-to-end method. The object detection can be achieved only by extracting features once, and the speed is faster than the two-stage algorithm, but the general accuracy is slightly lower. The application research of object detection in the field of construction engineering has a long period of exploration, and many outstanding research results have emerged in defect inspection [14], safety monitoring [25], and performance analysis [26].

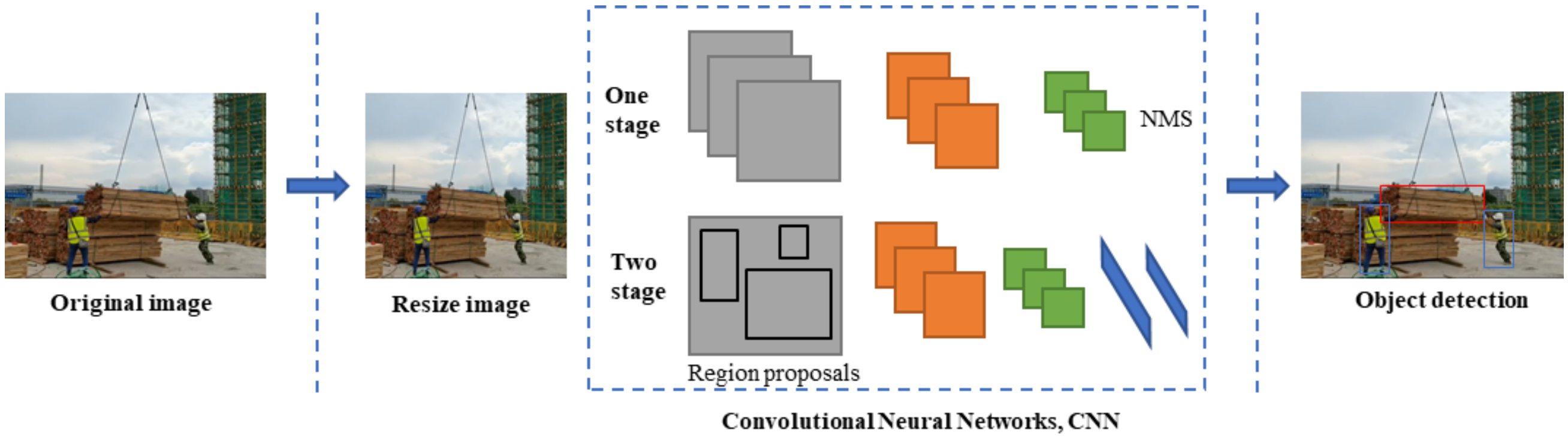

Fig. 1. Flowchart of the object detection algorithm

## 3. Methodology

In this study, we construct a new large-scale construction site image dataset, called Site Object Detection dAtaset (SODA), which contains 15 classes targets in four categories. We visited several construction sites in the city of Guangzhou, China, and collected more than 20000 images using different equipment, at different angles and times of the day. 35 students majoring in civil engineering were trained in image processing and annotation. Each student was responsible for about 600 images, which were then checked by several graduate students and experts in the AEC industry. As Figure 2 shows, the SODA building process mainly includes four steps: construction site category selection, image data acquisition, image data cleaning, image annotation.

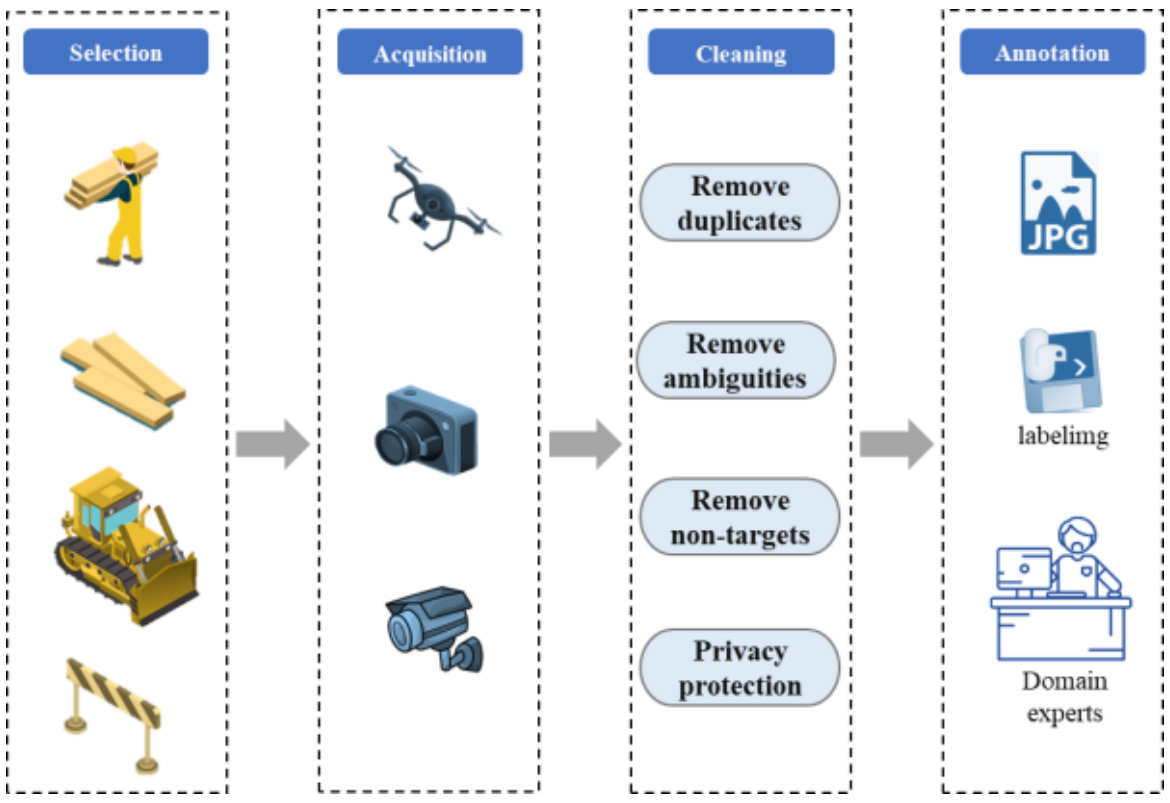

Fig. 2. The building process of SODA

### *3.1 Category Selection*

The common elements in the construction site can be classified into four categories: worker, material, machine, layout. We further expand 15 target detection classes suitable for deep learning object detection from these four categories. Worker categories include person, helmet, and reflective vests; material categories include board, wood, rebar, brick, and scaffold; machine categories include handcart, cutter, electric box, hopper, and hook. Layout categories include fence and slogan. Table 1 shows each of the types, and Figure 3 shows the example of the corresponding categories.

Table 1. Related description of selected objects in construction site

| **Construction site object** | **Description** | **example** |
|---|---|---|
| Person | Workers working in construction should wear PPE (safety hat, reflective vest). | Fig. 3(1) |
| vest | | Fig. 3(2) |
| helmet | | Fig. 3(3) |
| board | Board for construction engineering is used to support the weight and lateral pressure of concrete mixture with plastic flow properties, so that it can be solidified according to the design requirements. | Fig. 3(4) |
| wood | Wood into square shape according to the actual processing needs. It is generally used for decoration and door and window materials, template support, and roof truss materials in structural construction. | Fig. 3(5) |
| rebar | Rebar refers to the steel used for reinforced concrete and prestressed reinforced concrete. Its cross-section is circular, and sometimes it is a square with a round angle. | Fig. 3(6) |
| brick | Concrete block is a kind of lightweight porous, thermal insulation, good fire resistance, strong plasticity, and seismic capacity of new building materials. | Fig. 3(7) |
| scaffold | The scaffold is a working platform built to ensure the smooth progress of each construction process. | Fig. 3(8) |
| handcart | The handcart at the construction site is a two-wheeled, manual push and pull handling vehicle. | Fig. 3(9) |
| cutter | The cutter is the processing machine used in the material processing at the construction site. Commonly used machines are semi-automatic cutting machine cutting and CNC cutting machine cutting. | Fig. 3(10) |
| electric box | All electrical equipment in the construction site must have its own special electric switch box, which is convenient for the switching operation of the circuit and the reasonable distribution of electric energy. | Fig. 3(11) |
| hopper | Hopper (tower crane hopper, ash hopper, sand hopper, concrete hopper) are mainly used in building foundation, pouring concrete, piling, high-rise building construction material transportation. | Fig. 3(12) |
| hook | The hook of a tower crane is used to connect objects and ropes. | Fig. 3(13) |
| fence | The construction fence is the protective equipment to prevent accidental injury in the construction site of building engineering | Fig. 3(14) |
| slogan | The slogan of the construction site is used to alert workers to civilized and safe construction, spread enterprise culture, etc. | Fig. 3(15) |

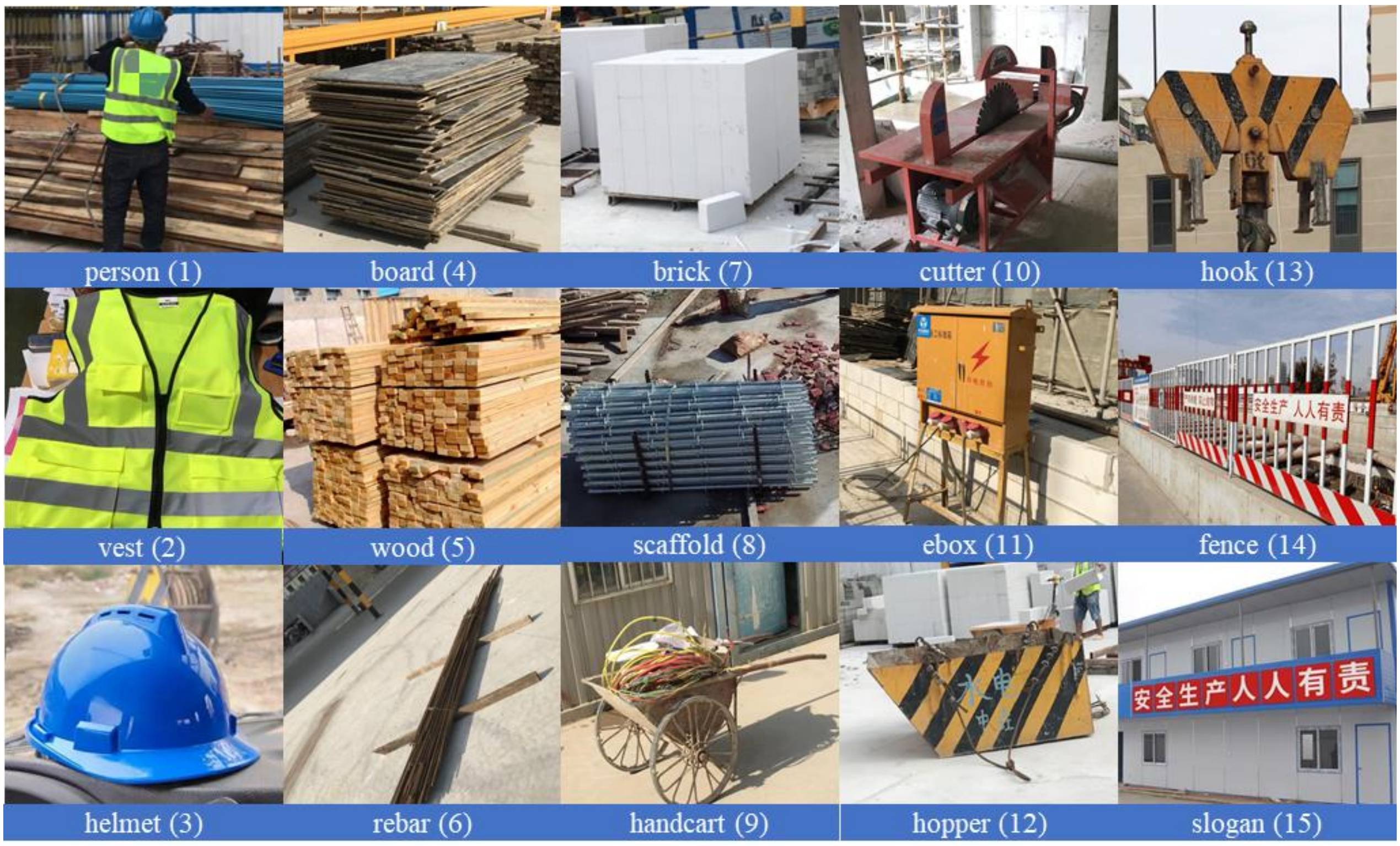

Fig. 3. Example of the corresponding categories.

Table 2. Categories and each class of the object label

| **category** | **label** | | | | |
|---|---|---|---|---|---|
| **Person** | person | helmet | vest | | |
| **Material** | board | wood | rebar | brick | scaffold |
| **Machine** | handcart | cutter | ebox | hopper | hook |
| **Layout** | fence | slogan | | | |

### *3.2 Data acquisition*

Conventional dataset construction often collects images from web search engines [41]. We have therefore initially tried to use web crawlers and other means to obtain data online, but the obtained images do not meet the requirements. Since the construction site is often messy, our objects are often mixed with many other unrelated objects, so our dataset are all collected on the real construction site. Field data acquisition mainly adopts three methods, UAV, handheld monocular camera shooting, and construction site monitoring video (hook visualization). All collected photos will eventually be converted into JPG format. We visited several construction sites covering various construction stages, from the foundation pit stage to the decoration stage. As a result, a total of 21863 images were collected.

In the process of data collection, we also found some noteworthy problems in data acquisition that may benefit people who wish to do the same. 1. In the actual situation of the construction sites, compared with workers and materials, the number of machines and layout is relatively small, so it is necessary to select a full range of shots of limited targets from multiple angles when shooting. 2. Due to the confusion, visual blind spots, and occlusion of construction sites, it is difficult to collect positive samples using a single shooting method, different shooting methods should be integrated. As shown in Figure 4, the hook is shot from different angles using different devices so as to provide different perspectives. Using UAV and handheld shooting on construction site also introduces security risks, where careful operation is needed. 3. The data obtained from the highest point camera can record the panoramic view of the construction site, and can also be photographed for large machines such as tower cranes. However, these video data are too vague to label small objects.

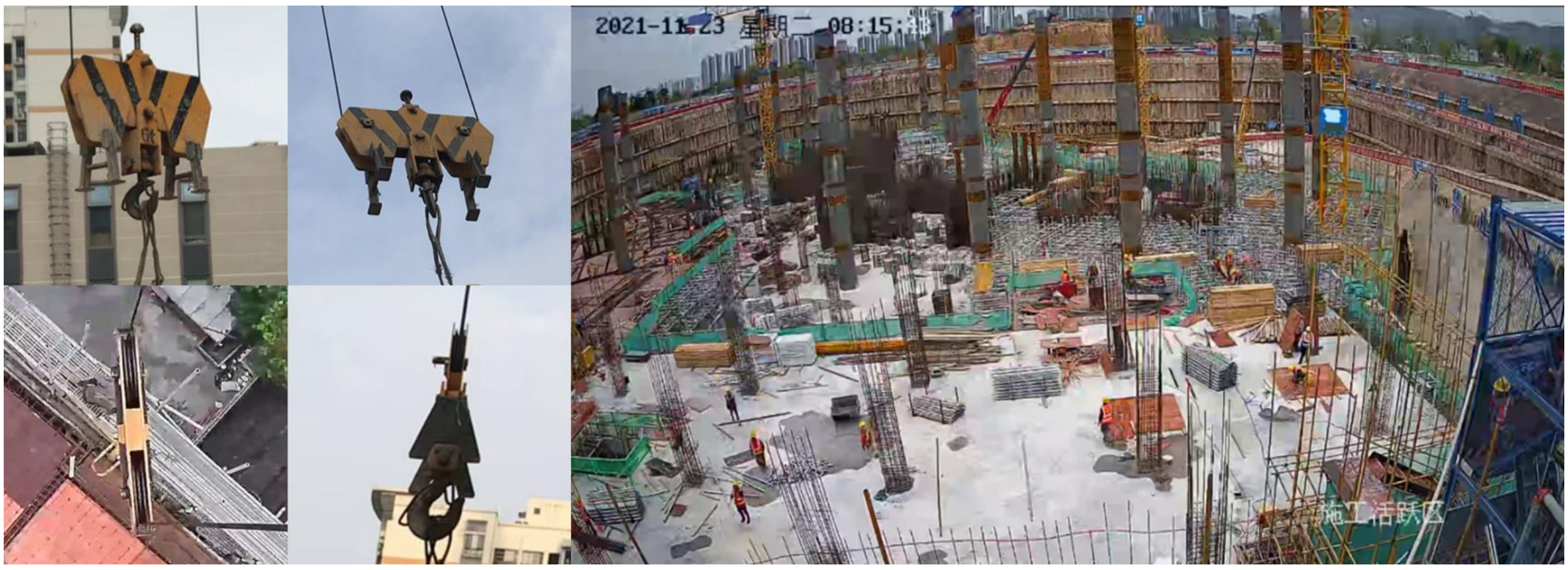

Fig. 4. Example of different angles using different devices.

### *3.3 Data cleaning*

After the completion of data acquisition, it cannot be annotated directly but should be preprocessed to eliminate invalid data. Four objectives of data cleaning are proposed: remove duplicates, remove ambiguities, remove non-targets, and corresponding privacy protection. The removal criteria and examples are shown in Figure 5.

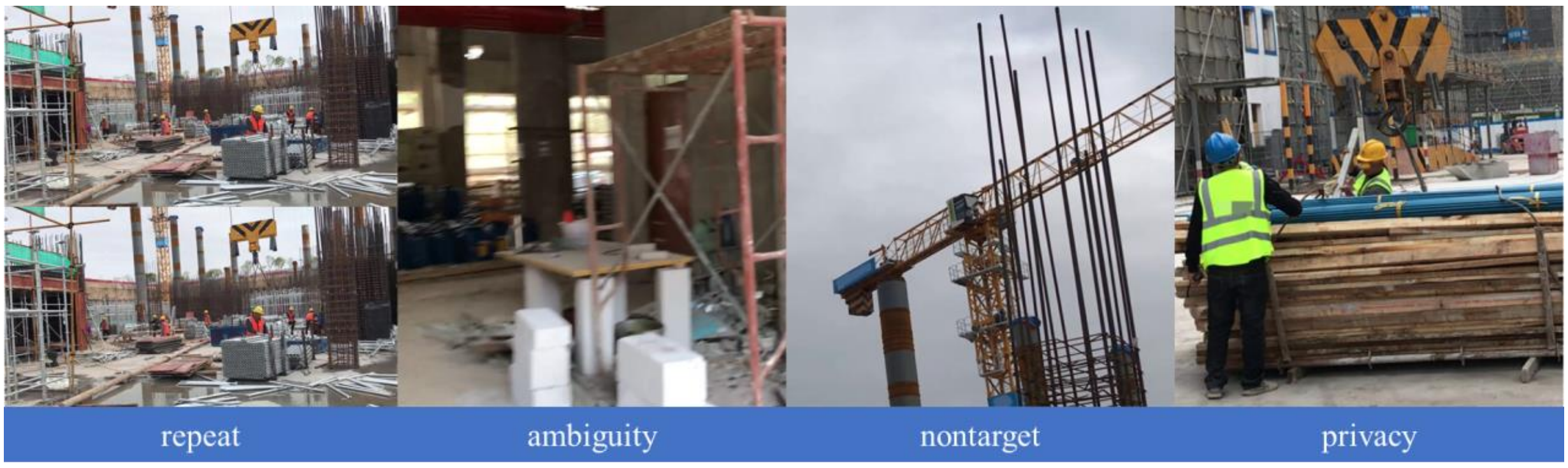

Fig. 5. Example of images that need to be removed and processed.

1. Remove duplicates

The production of the dataset requires that each image must be significantly different from other images in terms of angle, position, or illumination. Therefore, some repetitive images should be manually removed. In the shooting process of handheld monocular camera, many similar images may be collected by pressing the shutter several times. Moreover, some images of SODA are obtained by intercepting video frames. Although it is set to capture images from video every 30 frames, there are still relatively repetitive images that need to be manually removed due to slow progress in some construction activities.

2. Remove ambiguities

In the shooting process, due to weather, lighting, human and other reasons, some videos and images are blurred. These images are not only difficult to annotate, but also affect the training effect of the deep learning model. Our students delete them before or during annotation.

3. Remove non-targets

In the process of on-site shooting, it is inevitable to shoot some images that meet the conditions of non-repetition and non-ambiguity but do not contain the target of our dataset. It is of no significance for the research, and it also needs to be manually removed.

4. Privacy protection

Privacy is an important issue for publishing public datasets. Our team spent a lot of time and effort on privacy processing (200 working hours for each person in annotation, accounting for two-thirds of the data cleaning). Considering the engineering ethics, our privacy processing is divided into two parts, LOGO and other company information processing and human characteristics processing. In order to avoid infringing on the relevant company's commercial secrets or triggering related property rights issues, our data centralized processing company LOGO. At the same time, to avoid the ethical issues in the study, the face is fuzzified accordingly. The above specific operations were performed by manual processing privacy targets by the students.

The data cleaning process took about 300 hours, accounting for 25 % of the total dataset construction time. In the removal process, 2017 invalid images were deleted. In data privacy processing, after a round of privacy processing and a joint inspection by authors and experts, students still have 1307 omissions (37 omissions per person).

### *3.4 Data annotation*

The Visual Object Classes (VOC) format dataset is the standard dataset of the world-class Computer Vision Challenge (PASCAL VOC Challenge) [28], which is widely used in the field of computer vision and is an industry-recognized standard dataset format. In order to ensure the high quality of annotation, we adopt three standards when annotating an image, and strictly follow these three standards in the annotation process： 1. The annotation box must frame the target and does not intersect the target. 2. On the basis of 1, try to reduce the frame selection of irrelevant background on the basis of the framed target. 3. For similar targets with close distance, using a single target multiple frame selection without using a frame to select multiple targets. 4. When the target has partial occlusion or is inconvenient to annotate, they should be omitted. As shown in figure 6, we use the annotation standard of the blue box instead of the red box.

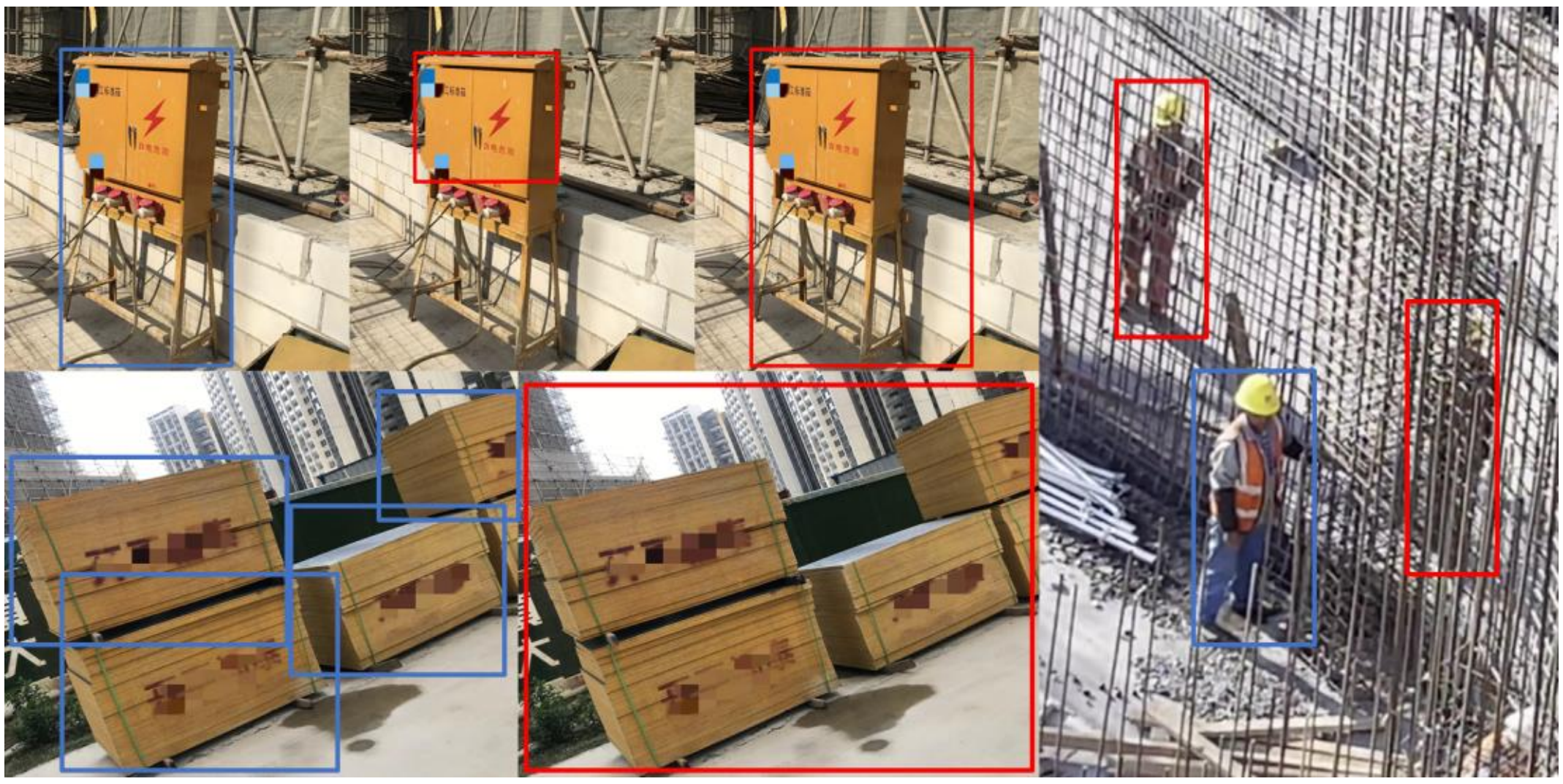

Fig. 6. Example of annotation standard.

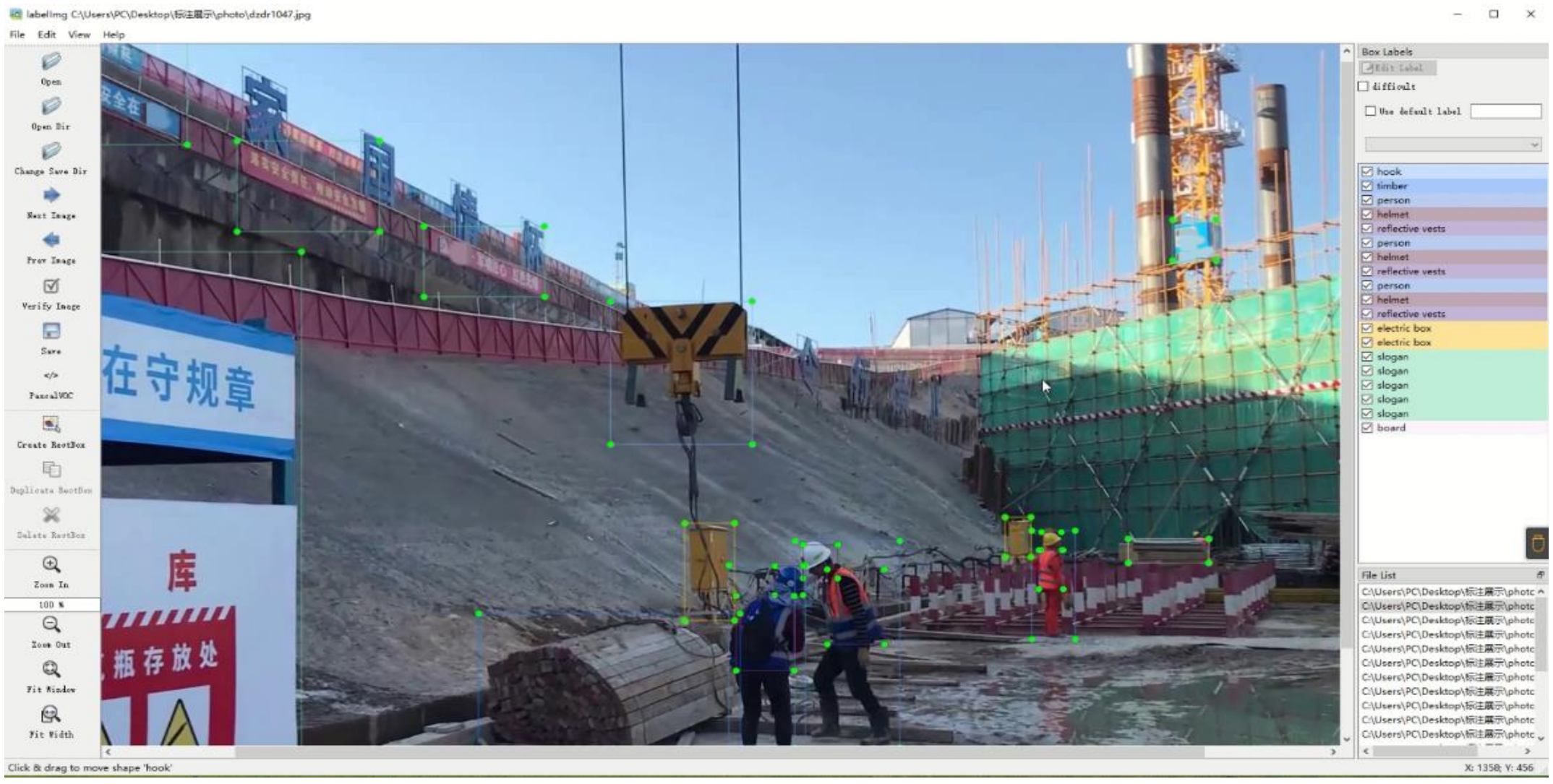

Fig. 7. The interface of software labelImg

As shown in the following Figure 7, students used software labelImg to annotate 15 categories of images and generated corresponding XML format files after labeling. The XML format is shown in Figure 8. It contains information such as storage path, image name, coordinate. Firstly, training 35 annotators and making tutorials to explain how to annotate and frame targets. The author regularly investigates the progress of annotation. In order to ensure the accuracy of the annotation, the label completed by the students will be examined jointly by a graduate student and the author.

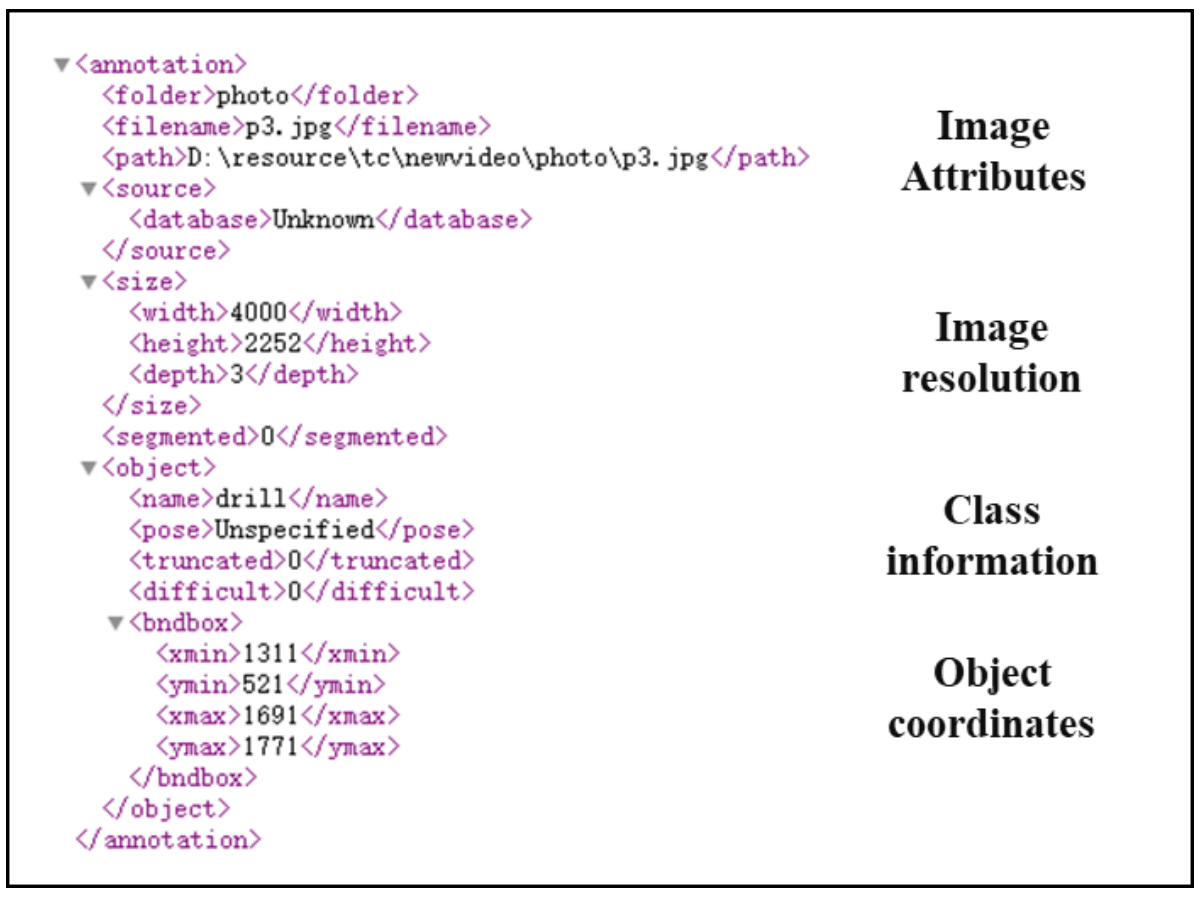

Fig. 8. XML format files

After obtaining the annotation documents of 35 students, the author conducted a round of inspections with another expert. Although they have been well trained, there are still some unexpected errors in the data. After data filtering, several categories outside our plan have been found. We found that the error concentrated in 1. The spelling error of tag words. 2. The plural error of the label. 3. Unknown labels. The label errors were listed as follows.

**Table 3. Example of the spelling mistake**

| | sample | | | | | | |
|---|---|---|---|---|---|---|---|
| **Right** | scaffold | hopper | helmet | board | person | vest | brick |
| **Error** | seaffold | hooper | helemt | borad/ boadrd | preson | vests | bricks |
| **Number** | 14 | 8 | 8 | 4/1 | 1 | 1 | 345 |

As Table 3 shows, all the word spelling errors and word plural errors were listed. In addition, there is another kind of error that 48 ‘w’ and 2 ‘wwww’ labels appear. The reason for the ‘w’ error is that the shortcut key of the tagging software is w, and the tag reporter typed the tag name incorrectly. All of these errors have been corrected in the officially released dataset.

After the above series of work, all students' data cleaning and annotation time were counted for 1246 hours, and the processing time was intermittent for nearly a month. We obtained the effective data of 19846 images and the corresponding xml files. Among them, data cleaning costs 300 working hours (account for 16.4%), data labeling costs 1246 working hours (account for 68%), and data checking costs 286 working hours (account for 15.6%).

## 4. Statistics of the dataset

In this part, we analyze our dataset and compare it with the AEC industry's current open-source object detection dataset. SODA contains a total of 19846 images, and the size of most images in the dataset is 1920 * 1080, accounting for 86%. A total of 286201 objects are included, and the object distribution is shown in Figure 10, which can provide a quantitative understanding of the dataset. Among them, the number of worker labels is the largest, and the labels of the machine and layout are less, which is more consistent with the data collection process on the construction sites. Each class has more than 1000 targets, and each category has more than 20,000 targets.

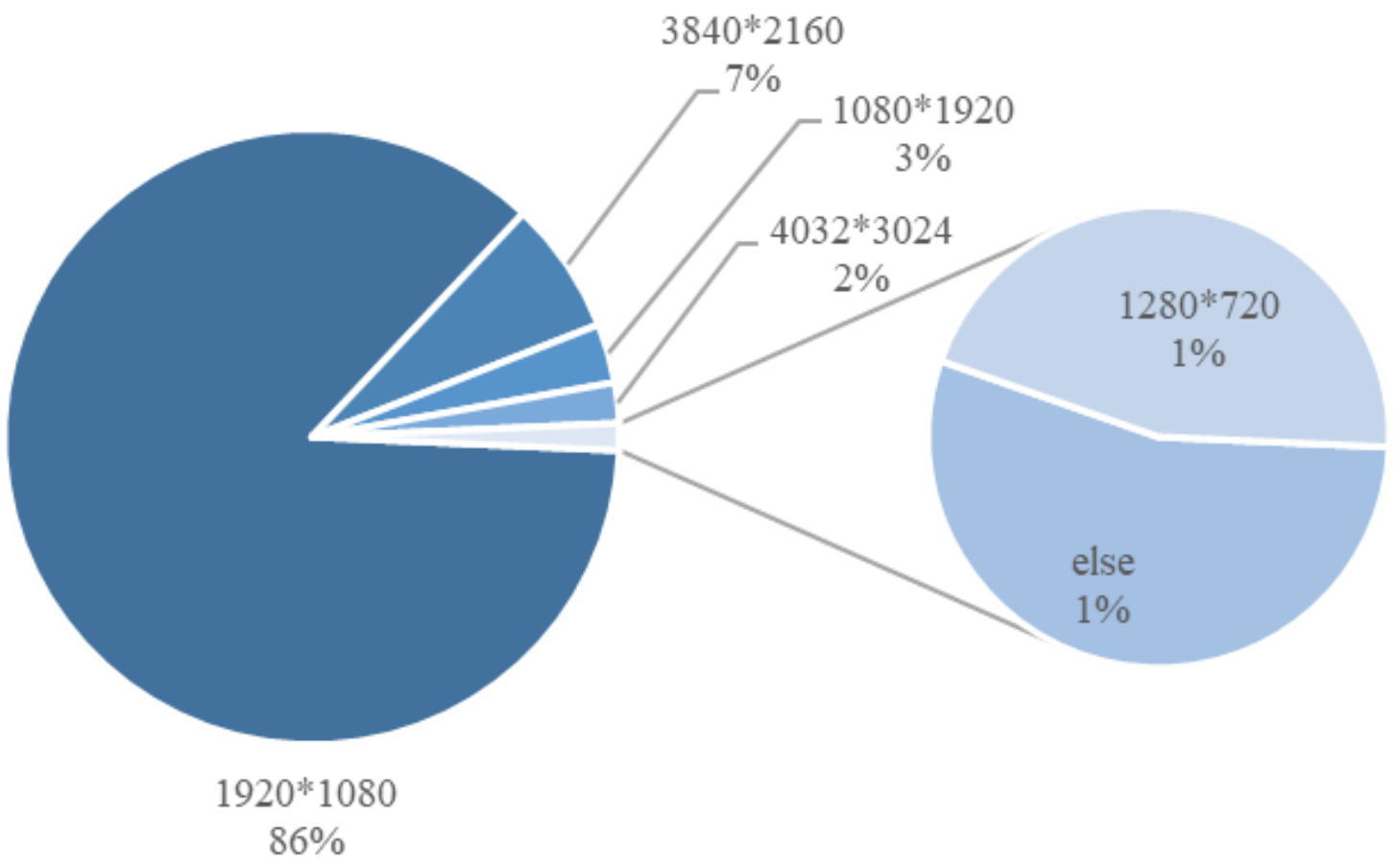

Fig. 9. Distribution of Image size by resolution

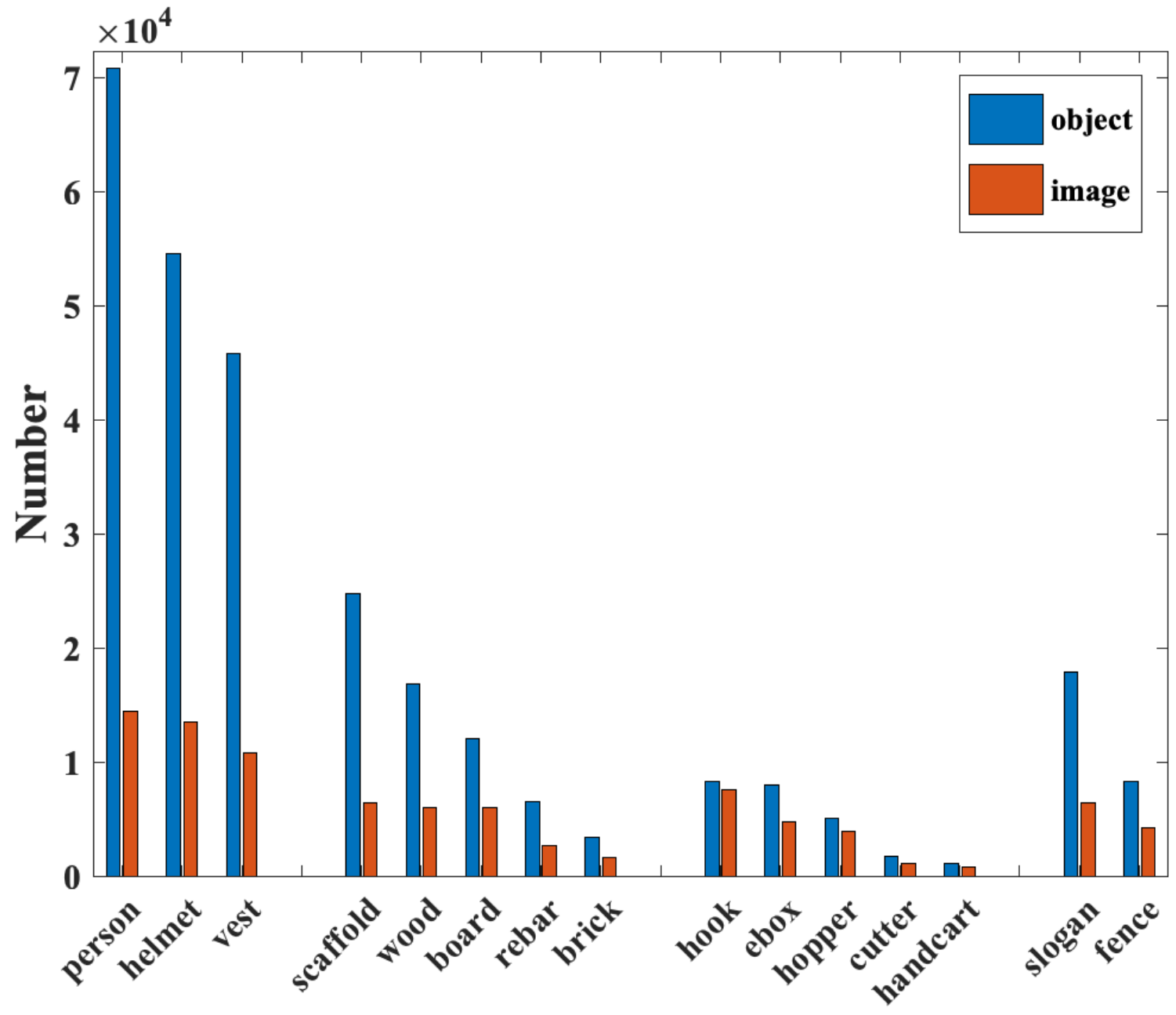

Fig. 10. Number of objects and images

Some deep learning object detection models generally specify the length-width ratio and range of the detection object, there is a statistical sample length-width clustering requirement for the dataset. The k-means clustering algorithm [42] is thus used to analyze the sample data. According to the sample data, the length-width ratio and range are visualized by clustering analysis. The results are shown in Figure 11.

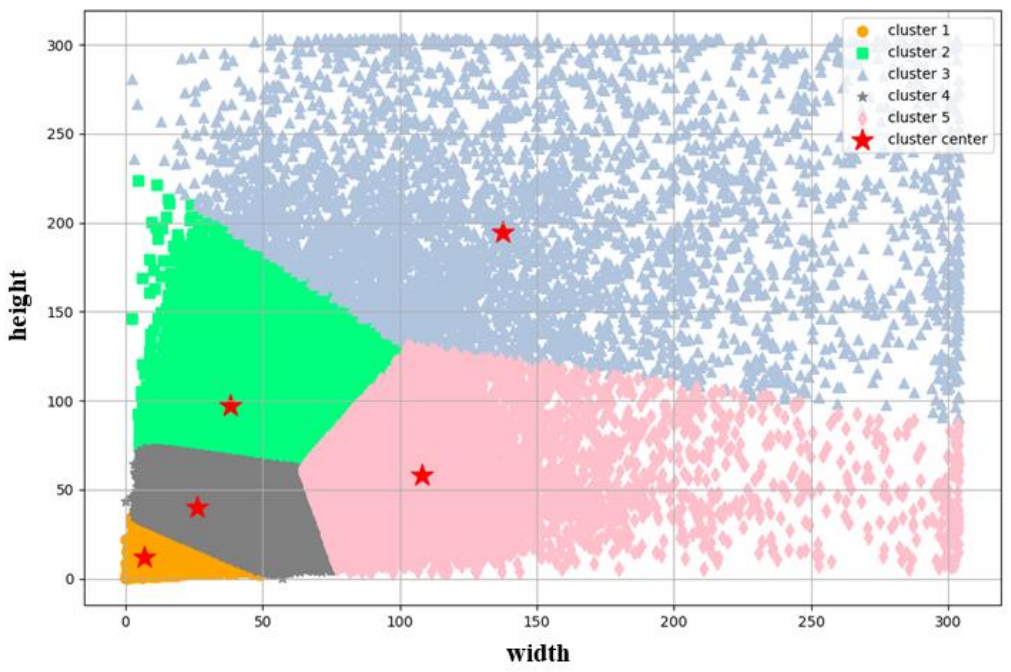

Fig. 11. Anchors statistical results of k-means clustering algorithm

It is also worth noting that we use different angles to take the pictures, with hand-held camera short-range shooting perspective, hand-held camera long-range shooting perspective, UAV perspective, tower crane hook visual system perspective, in order to achieve full coverage. The distribution is shown in Figure 12.

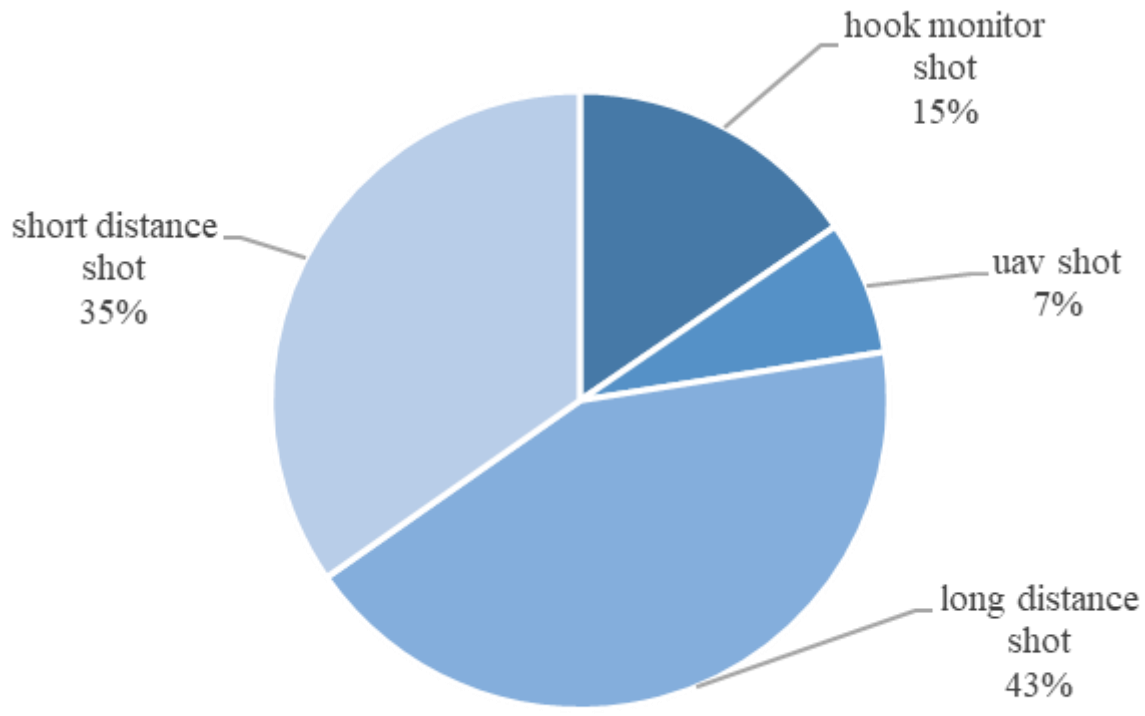

Fig. 12. Distribution of shooting angle in the dataset

Table 4 makes the comparison of our dataset with the current open-source object detection dataset in the construction industry. It is found that compared with the current image dataset dedicated to the construction industry, SODA not only contains the largest number of objects and categories but also is the first time to realize the full coverage of the four categories of workers, materials, machines, and layout. Moreover, we first used the image data of hook visualization equipment, which is also a neglected part of previous studies.

Table 4. Comparison of SODA dataset with other datasets of building construction

| Dataset | Image | Object | Description | Size |
|---|---|---|---|---|
| SODA | 19846 | 286201 | The 15 classes of objects in the construction site including four categories of workers, materials, machines, layout are annotated. | 1920*1080 and other |
| MOCS | 41668 | 222861 | The 13 classes of moving objects (worker and mobile machine) in construction sites are annotated. | 1200*-- |
| ACID | 10000 | 15,767 | Collect 10000 images of 10 construction machines and annotate machine types and their corresponding positions on the images. | >608*608 |
| CHV | 1330 | 9209 | Special dataset of 1330 images for multiple PPE classes considering actual construction site background. | 608*608 |

## 5. Experiments on the dataset

This study also aims to provide benchmarks for researchers to select appropriate algorithms for their subsequent applications. We also welcome other researchers to use different improved deep learning algorithms to verify our dataset and compare the results with the algorithm analysis results of this study. We selected two kinds of object detection algorithms to verify their performance on our dataset, which covered the one-stage object detection algorithm YOLO v3 and YOLO v4. The reason why YOLO is chosen is that it not only combines excellent methods such as deep residual network, feature pyramid, and multi-feature fusion network but also has higher detection speed and accuracy performance compared with the two-stage object detection model. The experiment was performed on a computer with the following configuration: NVIDIA Geforce RTX 2060, Intel (R) Core (TM) i7-10750H CPU @ 2.60GHz 2.59GHz 16.0GB. All algorithms are realized using the PyTorch framework in python language. SODA was divided into 90 % training set and 10 % test set and two object detection algorithms with configured parameters are selected to train the training set, and then the trained model is tested on the dataset.

We use the mean average precision (mAP) [43] to evaluate the quality of the model. The classification and localization of models in object detection tasks need to be evaluated, and each image may have different categories of different object targets. Therefore, the standard metrics used in image classification cannot be directly applied to object detection. The mAP gets rid of the limitation of using a single evaluation index by combining Precision and Recall.

Intersection over union (IOU) is a basic indicator for evaluating object detection algorithm performance, which is used to measure the coincidence degree between the detection box and the ground truth box. The calculation of IOU is shown in Figure 13. In the molecular part, the value is the overlap area between the detection box and the ground truth box. In the denominator part, the value is the total area occupied by the detection box and the ground truth box. After obtaining the IOU index, we need to calculate TP (True Positives), TN (True Negatives), FP (False Positives), FN (False Negatives). T or F represents whether the sample is correctly predicted. P or N represents whether the sample is predicted to be positive or negative. TP represents the positive sample is predicted correctly, TN represents the negative sample is predicted correctly, FP represents the negative sample is predicted wrongly, and FN represents the positive sample is predicted wrongly. After obtaining the four indicators, we can calculate Precision and Recall as follows.

Average precision (AP), using the combination of different Precision and Recall points, the area under the curve is obtained after integration. By calculating the AP of all classes, the Mean Average Precision (mAP) of the overall performance of the model can be estimated when detecting possible classes.

$$Precision = \frac{TP}{TP + FP}$$

$$Recall = \frac{TP}{TP + FN}$$

$$AP = \int_0^1 p(r)dr$$

$$mAP = \frac{1}{N}\sum_{n}^{i=1} AP$$

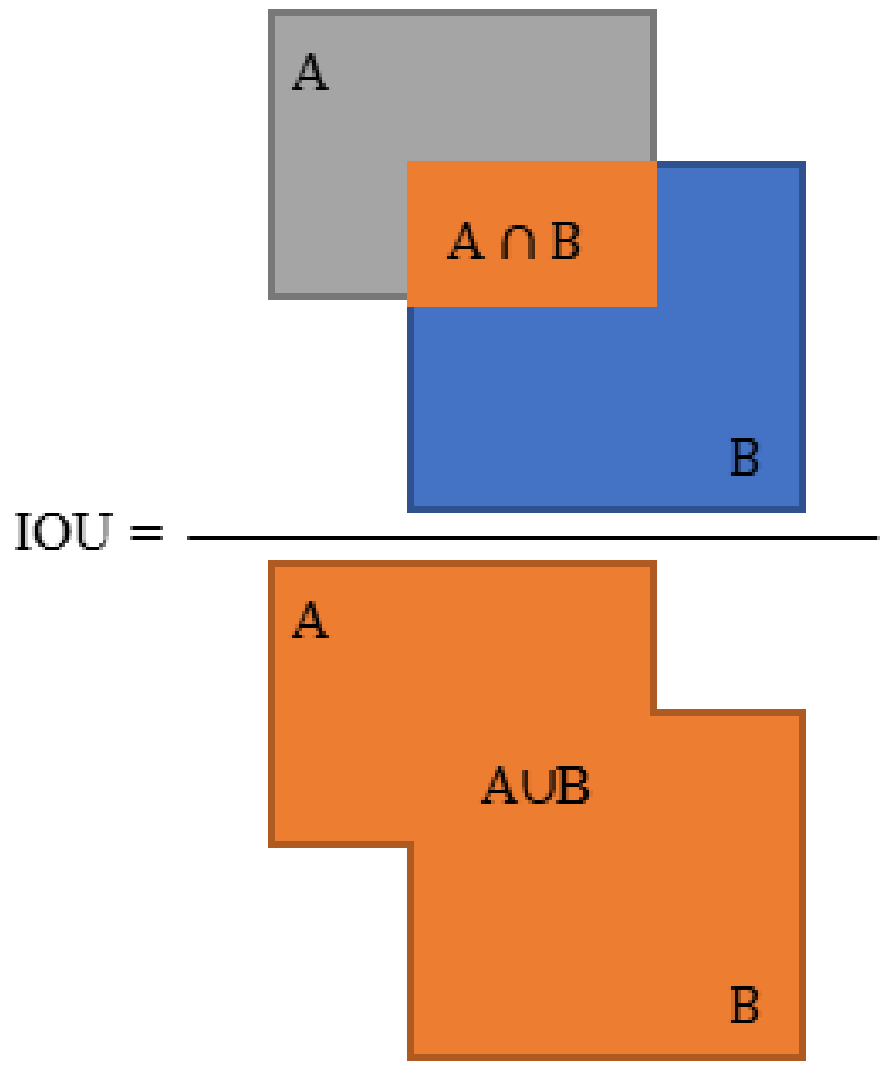

Fig. 13. The calculation process of the IOU

SODA was randomly divided into a training set (17861 images) and a test set (1985 images) according to 9:1. We selected and built the two main-stream deep learning algorithms to train and test datasets. And then the SODA was loaded for training. We adopted 100 epochs for training. Training is divided into two stages, namely the freezing stage and the thawing stage. The first 50 epochs freeze the main parameters of the model, increase the learning rate, help the model training jump out of the local optimal solution, and fine-tune the overall parameters. After 50 epochs of thawing model backbone parameters, reducing the learning rate, model backbone parameters in this stage were greatly changed.

The following Figure 14 is the training curve loss (training and verification loss) trained by two deep learning algorithms on SODA, in which the red line is the training loss and the blue line is the verification loss. It can be seen from the curve that the two deep learning algorithms are well fitted on the SODA dataset. It should be noted that the loss values of different algorithms are not necessarily comparable because the test algorithm implements different loss functions. In the training process, the training loss and verification loss continue to decrease with the increase of epoch, and at the same time, they decrease after the rise of the 50th thaw epoch, and the verification loss is higher than the training loss. The learning curve shows the robustness and universality of the dataset.

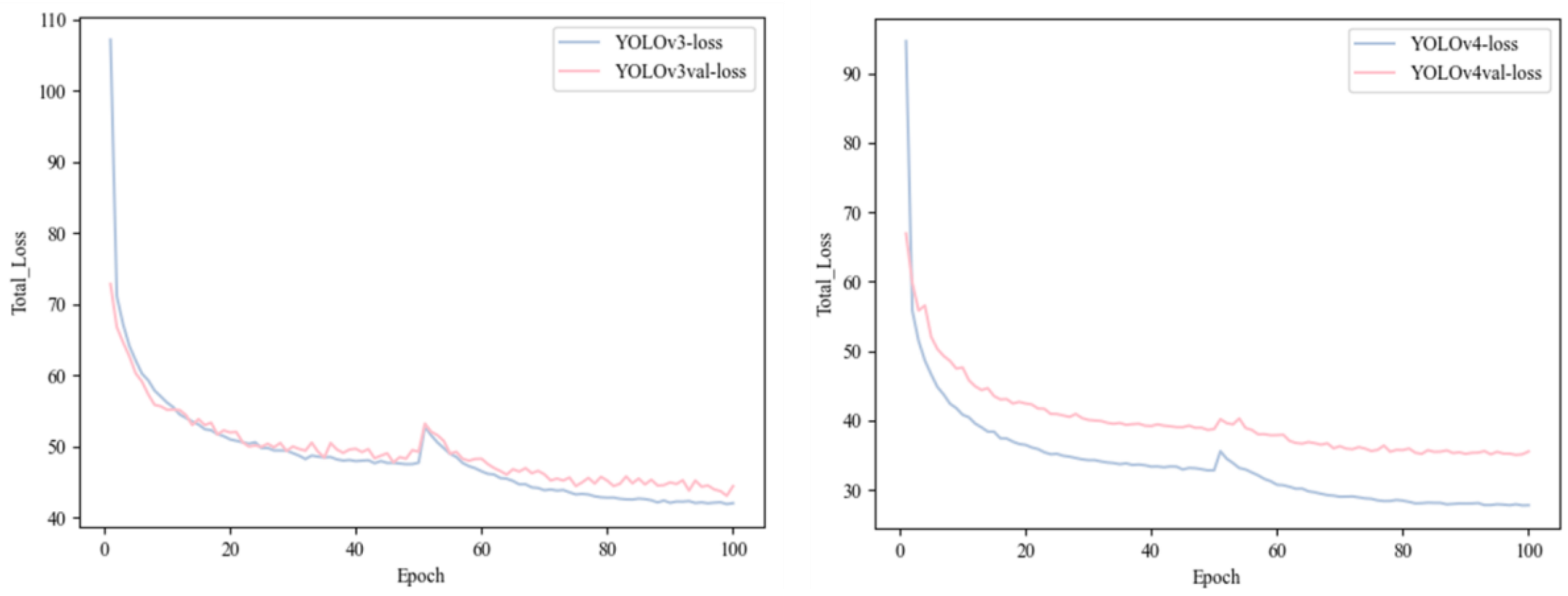

Fig. 14 The training loss and validation loss of YOLO (v3, v4)

After the training, the detector is evaluated on the test dataset. The overall performance index analysis results of the test dataset in YOLO v3 and YOLO v4 are shown in Table 5. The results show good detection performance. It has 71.22 % mAP on YOLO v3 and 81.47 % mAP on YOLO v4. Due to the change of network structure, each category has a different performance on the two networks, but it is worth noting that the training results of the two models show that the mAP of the material is higher, and the results of the workers are lower. The highest AP is the hook (92.81 % in YOLO v3) and the hopper (95.18 % in YOLO v4). The lowest AP is the fence (50.99 % in YOLO v3) and the helmet (55.62 % in YOLO v4). In terms of detection speed, YOLO v4 performed better at 31.94 FPS than 25.06 FPS of YOLO v3. Experimental results provide a benchmark for SODA.

In order to illustrate the practicability of research, the trained object detection model is applied to actual complex construction site images for verification. As shown in Figure 15 and Figure 16, it can be seen from the case that the YOLO v3 and YOLO v4 algorithms trained in this research have achieved good detection results. As shown in Figure 17, the identified classes are all correctly classified. The recognition performance of YOLO v3 and YOLO v4 is similar in simple scenes. But YOLO v4 can identify more categories and has higher recognition accuracy in complex scenes.

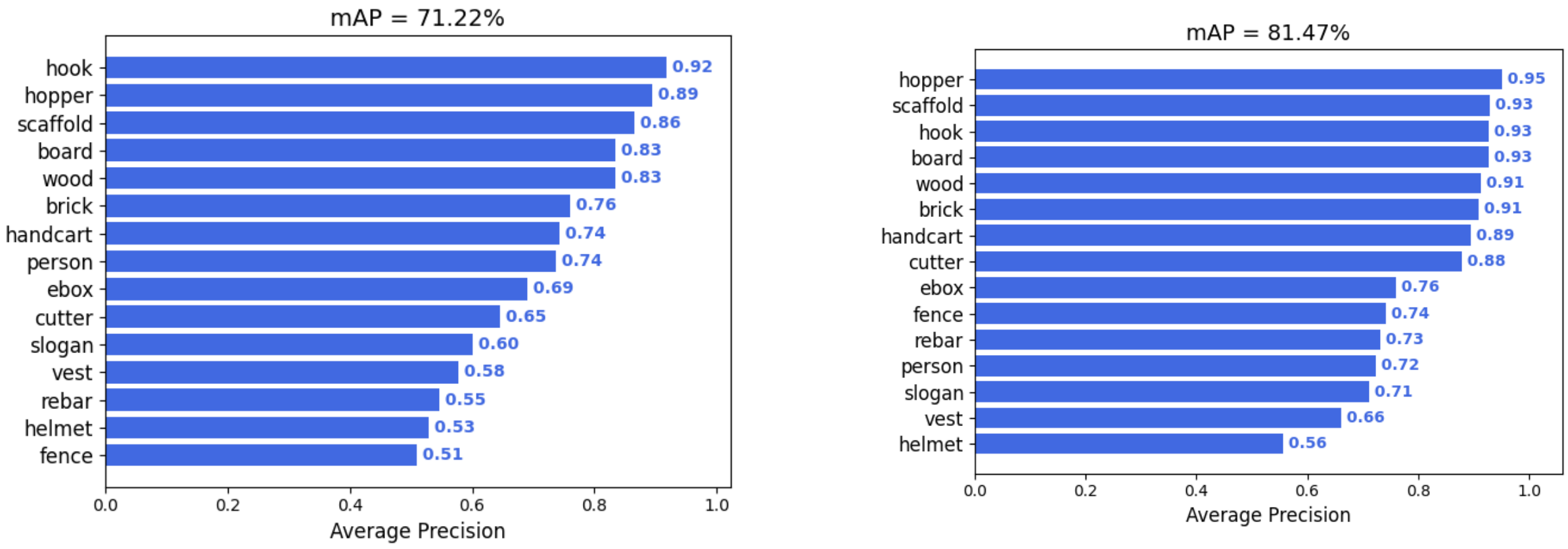

Fig. 15 The mAP of YOLO (v3, v4)

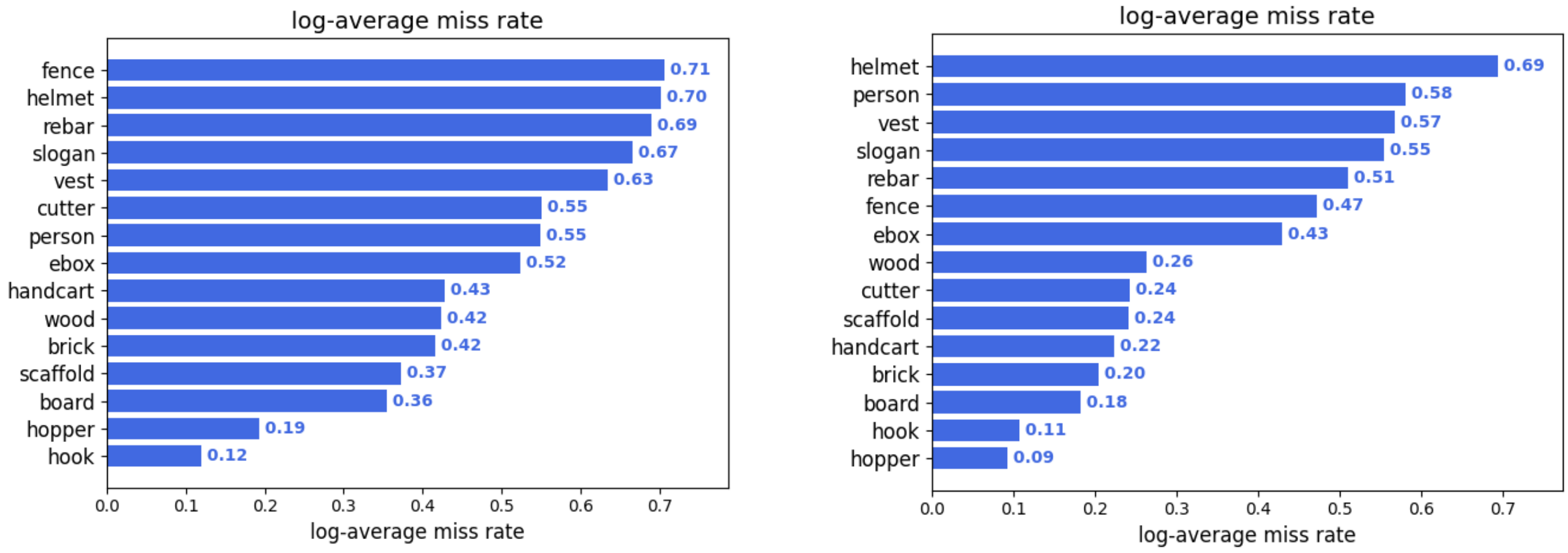

Fig. 16 The log-average miss rate of YOLO (v3, v4)

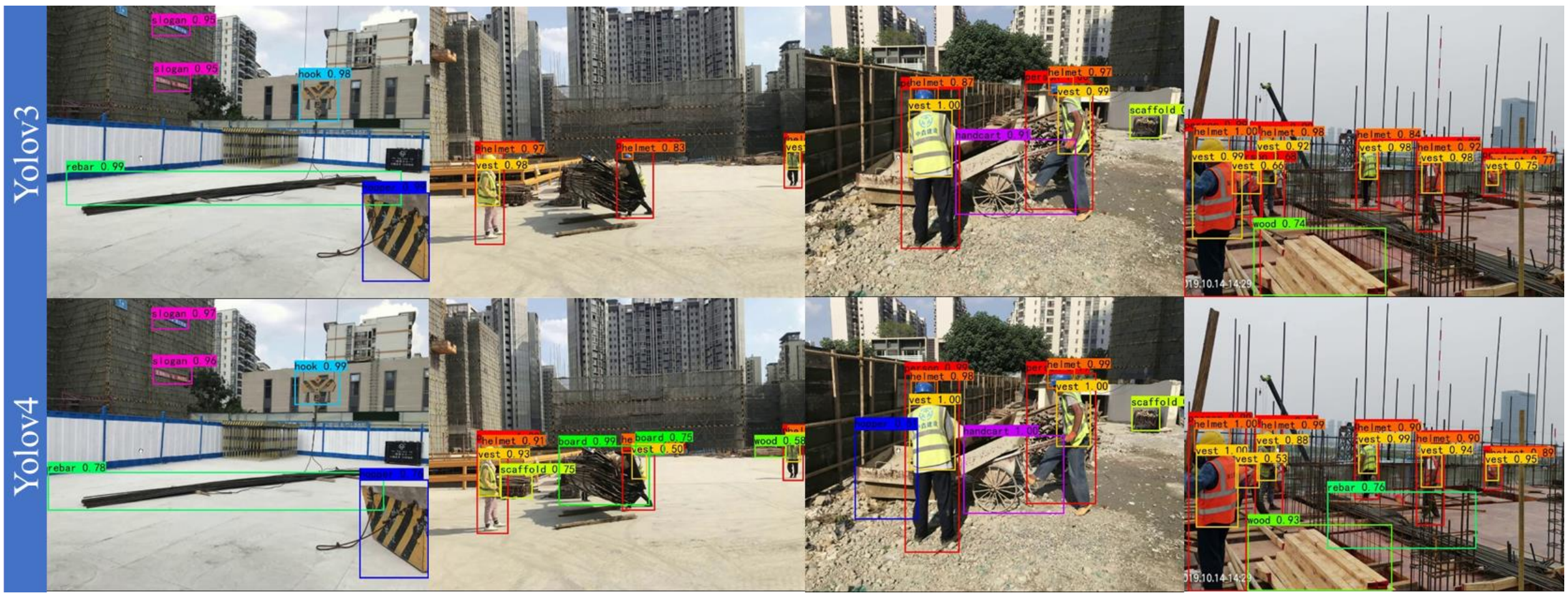

Fig. 17 The identification result of YOLO v3 and YOLO v4

Table 5. Performance comparison of SODA dataset in YOLO v3 and YOLO v4*

| **Algorithm** | **Worker (%)** | | | **Material (%)** | | | | | **Machine (%)** | | | | | **Layout (%)** | | **mAP** |
|---|---|---|---|---|---|---|---|---|---|---|---|---|---|---|---|---|
| | person | helmet | vest | board | wood | rebar | brick | scaffold | handcart | cutter | ebox | hopper | hook | fence | slogan | |
| YOLO v3 | 73.71 | 52.88 | 57.67 | 83.34 | 83.32 | 54.73 | 76.02 | 86.43 | 74.34 | 64.61 | 69.05 | 89.42 | **91.76** | **50.99** | 60.07 | 71.22 |
| YOLO v4 | 72.42 | **55.62** | 66.09 | 92.79 | 91.41 | 73.21 | 90.89 | 92.92 | 89.45 | 87.96 | 76.11 | **95.18** | 92.81 | 74.11 | 71.08 | 81.47 |

*The best performance and the worst performance of the model are roughened with red and green in the chart.

## 6. Conclusion and future work

In this research, we first created a SODA dataset for object detection in the construction site. The SODA dataset is an image dataset in VOC format, which contains 19846 images and annotation information. SODA includes 286201 objects, 15 categories, covering most of the common objectives of the construction site. Then, SODA is divided into a training set and a test set at a ratio of 9: 1, which are loaded into the mainstream one-stage object detection algorithms for training. A series of training results and benchmarks were obtained.

In summary, the contribution of SODA is as follows: 1. All the images are collected from actual construction sites at different construction stages, angles, and times of the day. 2. Achieve full perspective coverage shooting: near and far perspective, vertical perspective, UAV perspective, monitoring perspective; 3. Both quantity and quality of datasets are considerable and reliable. 4. The target species included are relatively extensive with the elements of the worker, material, machine, and layout. As we know, this is the first time that four elements of the worker, material, machine, and layout are detected simultaneously in a deep learning model, which can be used as the benchmark for future research. This study is the first open-source dataset to provide the largest coverage target of the construction site, and explains the construction process of the dataset, the problems encountered, and the solution experience in detail. At the same time, this study also establishes a test benchmark to provide a basis for other applications to select detectors. This work can provide help for construction site safety monitoring, construction progress analysis, pedestrian monitoring, emergency response, and civilized construction.

However, several limitations of research should be mentioned. 1. It is recommended to add more categories and more data to enrich the SODA dataset. Although the category and number of SODA are larger than other datasets in the industry, they are still smaller than other datasets in the deep learning community and should be increased later. 2. Further research on annotation task. In this study, the annotation of the SODA dataset is object-level, and only the boundary frame of the object is labeled, rather than pixel-level. This indicates that SODA can only be used to train deep learning object detection algorithms. The object segmentation algorithm that is more accurate than the bounding box cannot be trained. Pixel-level annotation dataset is another important work in the future. 3. The research focuses on the dataset construction process and does not carry out too much research on a deep learning algorithm. Only two classical deep learning models are trained. More classic models can be trained in the future to provide more benchmarks for the construction industry. 4. This dataset construction still uses manual annotation. Annotators are students majoring in engineering management. In the future, more annotation methods such as crowdsourcing annotation [44] and automatic annotation can be explored to construct datasets. 5. The data acquisition process of SODA relies on manual efforts. Although the quality of image data obtained by SODA is better than web crawlers, it is a time-consuming and labor-intensive process to collect suitable construction images and videos. Although many construction sites have installed surveillance cameras, the video captured is usually deleted after a fixed time or is difficult to obtain due to various other reasons. In the future, we will try to use access to the relevant authority in the construction industry to greatly enrich our dataset.

## Acknowledgments

The authors would like to acknowledge the support by Guangdong Science Foundation, Grant No. 2018A030310363, the support by the Science and Technology Program of Guangzhou, Grant No. 201804020069; Open Fund of Key Laboratory of Urban Land Resources Monitoring and Simulation, Ministry of Natural Resources (KF-2019-04-024); and the support by the State Key Lab of Subtropical Building Science, South China University of Technology (No. 2022ZB19).

The author would also like to pay special tribute to students who contribute to the data cleaning and annotation process of SODA at the South China University of Technology. Their names are Junxiong Zhang, Fengning Chen, Hongfeng Chen, Jianhe Chen, Jingjun Chen, Zhentao Chen, Yina E, Jie Fan, Xingyu Gao, Jiaxuan He, Jiayi Huang, Jingyuan Huang, Ying Huang, Yuefan Huang, Jiaxi Jiang, Liki Lei, Jufang Lin, Rui Liu, Junjie Ma, Yinchao Qiu, Wanxi Su, Ying Sun, Jiaquan Wang, Xinyuan Wang, Jide Wu, Haopeng Yan, Yuqi Zeng, Aiwaner Zeng, Xiaolan Jan, Yang Zhang, Honglong Zheng, Yuxian Zhu, Junze Zheng, Zhu Chao, and Yelin Ru.

## Data Availability

For pictures and annotations of SODA, start an immediate download from this link: https://scut-scet-academic.oss-cn-guangzhou.aliyuncs.com/SODA/2022.2/VOCv1.zip. Data regarding the process of the study can be reasonably obtained from the corresponding author.

## Credit authorship contribution statement

Rui Duan: Methodology (lead), Formal analysis, Investigation, Writing – original draft. Hui Deng: Resources, Validation, Writing – review & editing, Funding acquisition. Mao Tian: Methodology, Validation, Writing – review & editing. Yichuan Deng: Conceptualization (lead), Methodology, Writing – review & editing, Funding Acquisition. Jiarui Lin: Conceptualization, Resources, Validation, Writing – review & editing.